\documentclass[conference]{IEEEtran}
\IEEEoverridecommandlockouts
 
\usepackage{cite}
\usepackage{amsmath,amssymb,amsfonts}
\usepackage{algorithmic}
\usepackage{graphicx}
\usepackage{textcomp}
\usepackage{floatrow}
\newcommand{\genesys}{\textsc{GeneSys}\xspace}  
\newcommand{\adam}{\textsc{ADAM}\xspace}
\newcommand{\eve}{\textsc{EvE}\xspace}

\usepackage{amsmath}
\usepackage{algorithmic}
\usepackage[linesnumbered]{algorithm2e}
\usepackage{color}

\usepackage[hyphens]{url}
\usepackage{breakurl}
\usepackage{hyperref}

\usepackage[usenames,dvipsnames]{xcolor}
\hypersetup{citecolor=blue,linkcolor=blue}

\usepackage{graphicx}
\usepackage{listings}
\usepackage{setspace}
\usepackage{caption}
\usepackage{comment}
\usepackage{floatrow}
\usepackage{subfigure}
\usepackage[most]{tcolorbox}

\usepackage{endnotes,microtype,xspace,graphicx,fancyvrb,multirow}
\usepackage{fancyhdr}
\usepackage[normalem]{ulem}
\usepackage[hyphens]{url}

\newcommand{\system}{\textsc{Genesys}\xspace}

\newcommand{\subparagraph}{}
\usepackage[compact]{titlesec}
\titleformat{\paragraph}[runin]{\vspace{-3pt}\normalfont\normalsize\bfseries\small\sffamily
  }{}
  {0pt}{}[.]
  
\newfloatcommand{capbtabbox}{table}[][\FBwidth]
  
\titleformat{\subparagraph}[runin]{\normalfont\normalsize\itshape
 }{}
 {0pt}{}{}

\newcommand{\todo}[1]{\textcolor{red}{TODO: #1}}

\newcommand{\insertFigure}[2]{
    \begin{figure}[t]
\setlength{\abovecaptionskip}{0pt}
\setlength{\belowcaptionskip}{0pt}
        \centering
        \includegraphics[width=\linewidth]{figs/#1.pdf}
        \caption{\small #2}
        \label{fig:#1}
    \end{figure}
}

\newcommand{\insertNarrowFigure}[2]{
    \begin{figure}[t]
\setlength{\abovecaptionskip}{0pt}
\setlength{\belowcaptionskip}{0pt}
        \centering
        \includegraphics[width=0.8\linewidth]{figs/#1.pdf}
        \caption{\small #2}
        \label{fig:#1}
    \end{figure}
}

\newcommand{\insertWideFigure}[2]{

    \begin{figure*}[h]
\setlength{\abovecaptionskip}{0pt}
\setlength{\belowcaptionskip}{-3pt}
        \centering
        \includegraphics[width=\textwidth]{figs/#1.pdf}
	\vspace{-2mm}
        \caption{\small #2}
	\vspace{-4mm}
        \label{fig:#1}
    \end{figure*}

}

\newcommand{\squishlist}{
 \begin{list}{$\bullet$}
  { \setlength{\itemsep}{0pt}
     \setlength{\parsep}{3pt}
     \setlength{\topsep}{3pt}
     \setlength{\partopsep}{0pt}
     \setlength{\leftmargin}{1.5em}
     \setlength{\labelwidth}{1em}
     \setlength{\labelsep}{0.5em} } }

\newcommand{\squishlisttwo}{
 \begin{list}{$\bullet$}
  { \setlength{\itemsep}{0pt}
     \setlength{\parsep}{0pt}
    \setlength{\topsep}{0pt}
    \setlength{\partopsep}{0pt}
    \setlength{\leftmargin}{2em}
    \setlength{\labelwidth}{1.5em}
    \setlength{\labelsep}{0.5em} } }

\newcommand{\squishend}{
  \end{list}  }

\begin{document}
\title{GeneSys: Enabling Continuous Learning through Neural Network Evolution in Hardware\\
\thanks{This work is supported by NSF CRII Grant 1755876}
}

\author{\IEEEauthorblockN{Ananda Samajdar}
\IEEEauthorblockA{
\textit{Georgia Tech}\\
Atlanta, GA, USA \\
anandsamajdar@gatech.edu}
\and
\IEEEauthorblockN{Parth Mannan}
\IEEEauthorblockA{
\textit{Georgia Tech}\\
Atlanta, GA, USA \\
parth.mannan@gatech.edu}
\and
\IEEEauthorblockN{Kartikay Garg}
\IEEEauthorblockA{
\textit{Georgia Tech}\\
Atlanta, GA, USA \\
kgarg40@gatech.edu}
\and
\IEEEauthorblockN{Tushar Krishna}
\IEEEauthorblockA{
\textit{Georgia Tech}\\
Atlanta, GA, USA \\
tushar@ece.gatech.edu}
}
\maketitle

\begin{abstract}
Modern deep learning systems rely on (a) a hand-tuned neural network topology, (b)
massive amounts of labelled training data, and (c) extensive training over large-scale compute resources 
to build a system that can perform efficient image classification or speech recognition.
Unfortunately, we are still far away from implementing adaptive general purpose intelligent systems which 
would need to learn autonomously in unknown environments and may not have access to some or any of these three components.
Reinforcement learning and evolutionary algorithm (EA) based methods circumvent this problem by continuously interacting 
with the environment and updating the models based on obtained rewards.
However, deploying these algorithms on ubiquitous autonomous agents at the edge (robots/drones) demands extremely high energy-efficiency due to
(i) tight power and energy budgets,
(ii) continuous / life-long interaction with the environment,
(iii) intermittent or no connectivity to the cloud to offload
 heavy-weight processing.

To address this need, we present \system, a HW-SW prototype of a EA-based learning system, that comprises of a closed loop learning engine called EvE and an inference engine called ADAM.
EvE can evolve the topology and weights of neural networks completely in hardware for the task at hand, without requiring hand-optimization or backpropogation training.
ADAM continuously interacts with the environment and is optimized for efficiently running the irregular neural networks generated by EvE.
\system identifies and leverages multiple unique avenues of parallelism unique to EAs that we term ``gene"-level parallelism, and ``population"-level parallelism.
We ran \system with a suite of environments from OpenAI gym and observed 2-5 orders of magnitude higher energy-efficiency over state-of-the-art embedded and desktop CPU and GPU systems.
\end{abstract}

\section{Introduction}
\label{sec:intro}
\vspace{-2mm}

Ever since modern computers were invented, the dream of creating an intelligent entity 
has captivated humanity.
We are fortunate to live in an era when, thanks to deep learning, computer programs have paralleled, or in some cases even 
surpassed human level accuracy in tasks like visual perception or speech synthesis.
However, in reality, despite being equipped with powerful algorithms and computers, we are still far away from realizing general purpose AI.

The problem lies in the fact that the development of supervised learning based solutions is mostly open loop (\autoref{fig:learning_engine}(a)). 
A typical deep learning model is created by hand-tuning the neural network (NN) topology by a team of experts over multiple iterations, often by trial and error. 
The said topology is then trained over gargantuan amounts of labeled data, often in the order of petabytes, 
over weeks at a time on high end computing clusters, to obtain a set of weights.
The trained model hence obtained is then deployed in the cloud or at the edge over inference accelerators (such as GPUs, FPGAs, or ASICs).
Unfortunately, supervised learning as it operates today breaks if one or more of the following occur:\\
(i) unavailability of structured labeled data\\
(ii) unknown NN topology for the problem\\
(iii) dynamically changing nature of the problem\\
(iv) unavailability of large computing clusters for training.

\insertFigure{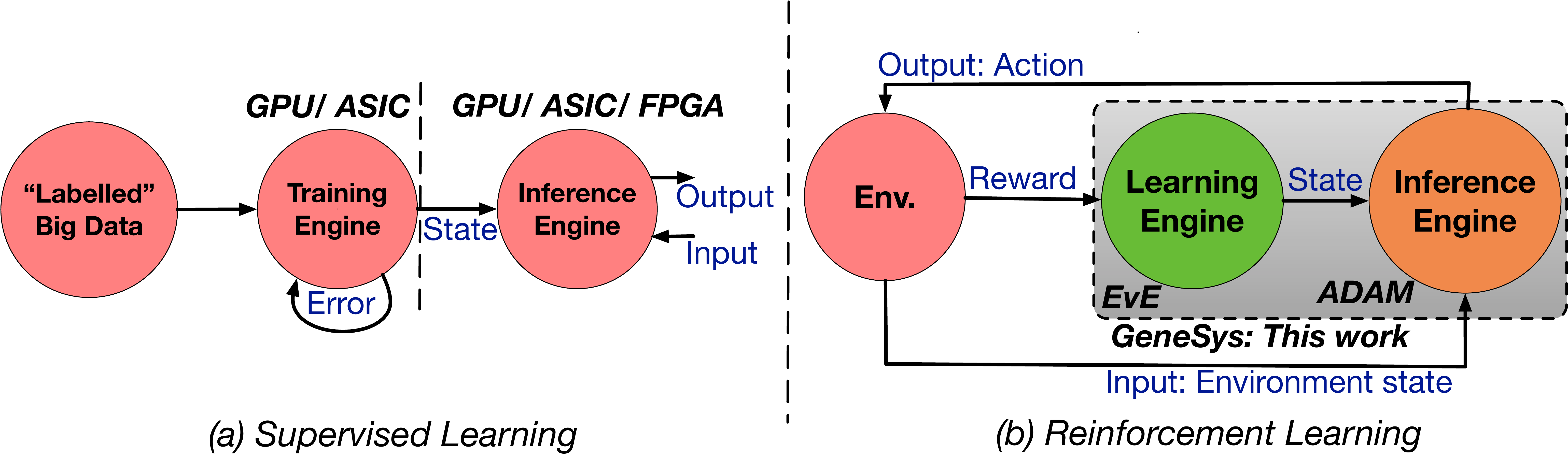}{Conceptual view of \system within machine learning.}
\insertFigure{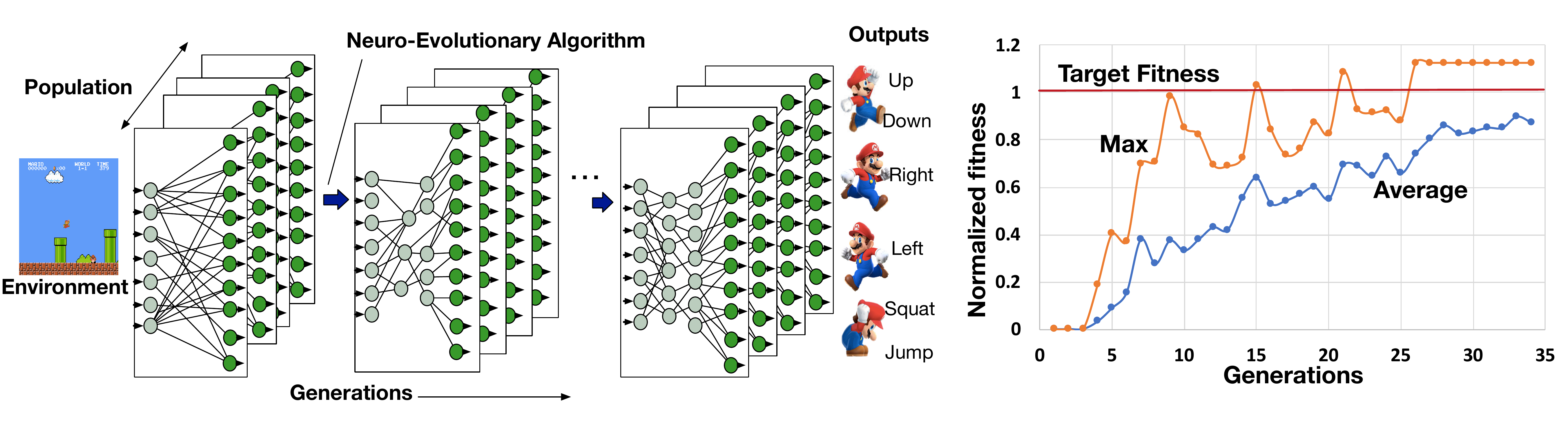}{Example of NE in action, evolving NNs to play Mario.}

At the algorithmic-level, there has been promising work on reinforcement learning (RL) algorithms, 
as one possible solution to address the first three challenges.
RL algorithms follow the model shown in \autoref{fig:learning_engine}(b).
An agent (or set of agents) interacts with the environment by performing a set of actions,
 determined by a 
policy function, often a NN.
These interactions periodically generate a reward value, which is a measure of effectiveness of the action for the given task. 
The algorithm uses reward values obtained in each iteration to update the policy function. This goes on till it converges upon the optimal policy.
There have been extremely promising demonstrations of RL~\cite{alphago, atari} - the most famous being Google DeepMind's
supercomputer autonomously learning how to play AlphaGo and beating the human champion~\cite{alphago}.
Unfortunately, RL algorithms cannot address the fourth challenge, for the same reason that supervised training algorithms cannot - 
they require backpropogation to train the NN upon the receipt of each reward which is extremely computation and memory heavy.

Bringing general-purpose AI to autonomous edge devices requires a co-design of the algorithm and architecture to synergistically solve all four 
challenges listed above.
This work attempts to do that.
We present \system, a system targeted towards energy-efficient acceleration of {\it neuro-evolutionary (NE) algorithms}.
NE algorithms are akin to RL algorithms, but attempt to ``evolve" the topology and weights of a NN via genetic algorithms, 
as shown in \autoref{fig:mario}.
NEs show surprisingly high robustness against the first 3 challenges mentioned earlier, and have seen a resurgence 
over the past year through work by OpenAI~\cite{salimans2017evolution}, Google Brain~\cite{real2017largescale} and Uber AI Labs~\cite{deepnevol}.
However, these demonstrations have still relied on 
big compute and memory (challenge \#4), which we attempt to solve in this work via clever HW-SW co-design.
We make the following contributions:
\squishlist
\item We characterize a NE algorithm called NEAT~\cite{neat}, 
identifying the compute and memory requirements across a suite of environments from OpenAI gym~\cite{openai}.
\item We identify opportunities for parallelism (population-level parallelism or PLP 
and gene-level parallelism or GLP) and data reuse (genome-level reuse or GLR) unique to NE algorithms, providing architects with insights 
on designing efficient systems for running such algorithms.
\item We discuss the key attributes of compute and communication within NE algorithms that makes them inefficient to run on GPUs and other DNN accelerators. 
We design two novel accelerators, \textsc{Evolution Engine (EvE)} and \textsc{Accelerator for Dense Addition \& Multiplication (ADAM)}, 
optimized for running the learning and inference of NE 
respectively in hardware, presenting architectural trade-offs along the way. \autoref{fig:learning_engine}(b) shows an overview.
\item We build a \system SoC in 15nm, and evaluate it against optimized NE implementations over latest embedded and desktop CPUs and GPUs. 
We observe 2-5 orders of magnitude improvement in runtime and energy-efficiency.
\squishend

Just like optimized hardware ushered in the Deep Learning revolution, we believe that \system and subsequent follow on work 
can enable mass deployment of intelligent devices at the edge capable of learning autonomously.
\section{Background}
\label{sec:background}
\vspace{-2mm}
Before we start with the description of our work, we would like to give a brief introduction to some concepts which we hope will help the reader to appreciate the following discussion.


\insertWideFigure{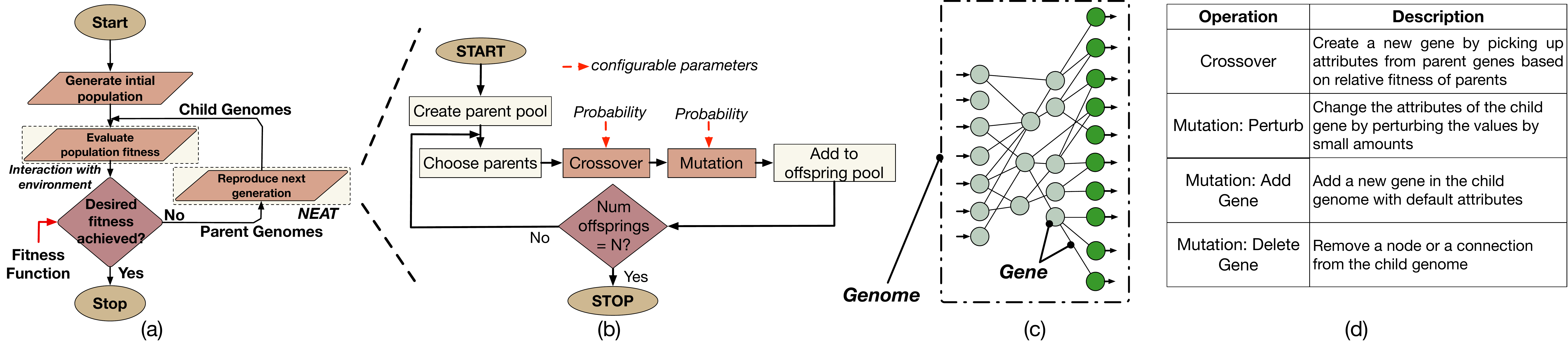}{(a) Flow chart of an evolutionary algorithm (b) The NEAT algorithm (c) Genes \& Genomes in context of NNs (d) Ops in NEAT. }

\subsection{Supervised Learning}
\label{subsec:supervise}
\vspace{-1mm}
Supervised learning is arguably the most widely used learning method used at present.
It involves creating a `policy function" (e.g., a NN topology) (via a process of trial and error by ML researchers)
and then running it through tremendous amounts of labelled data. 
The output of the model is computed for a given set of inputs and compared against an existing label to generate an error value.
This error is then backpropogated~\cite{backprop} (BP) via the NN to compute error gradients and update weights. 
This is done iteratively till convergence is achieved.


Supervised learning has the following limitations as the learning/training engine for general purpose AI:
\squishlist
\item{Dependence on large structured \& labeled datasets to perform effectively without overfitting ~\cite{imagenet, pascalvoc}}
\item {Effectiveness is heavily tied to the NN topology, as we witnessed with deeper {\it convolution} topologies~\cite{alexnet, resnet} that led to the birth of Deep Learning.}
\item {Extreme compute and memory requirements~\cite{dnn_training_sc2017, vdnn_micro2016}. It often takes weeks to train a deep network on a compute cluster consisting of several high end GPUs.}
\squishend

\subsection{Reinforcement Learning (RL)}
\label{subsec:reinforce}
\vspace{-1mm}
Reinforcement learning is used when the structure of the underlying policy function is not known.
For instance, suppose we have a a robot learning how to walk. The system has a finite set of outputs (say which leg to move when and in what direction),
and the aim is to learn the right policy function so that the robot moves closer to its target destination.
Starting with some random initialization, the agent performs a set of actions, and receives an reward from the environment for each of them,
which is a metric for success or failure for the given goal.
The goal of the RL algorithm is to update its policy such that future reward could be maximized.
This is done by iteratively perturbing the actions and computing the corresponding update to the NN parameters via BP.
RL algorithms can learn in environments with scarce datasets and without any assumption on the underlying NN topology,
but the reliance on BP makes them computationally very expensive.

\subsection{Evolutionary Algorithms (EA)}
\vspace{-1mm}
Evolutionary algorithms get their name from biological evolution, since at an abstract level
they be seen as sampling a population of individuals and allowing the successful individuals to determine the constitution of future generations.
\autoref{fig:evol_algo}(a) illustrates the flow.
The algorithm starts with a pool of individuals/agents, each one of which independently tries to perform some action on the environment to solve the problem.
Each individual is then assigned a fitness value, depending upon the effectiveness of the action(s) taken by them.
Similar to biological systems, each individual is called a {\it genome}, and is represented by a list of parameters called {\it genes} that each encode a particular characteristic of the individual.
After the fitness calculation is done for all, next generation of individuals are created by crossing over and mutating the genomes of the parents.
This step is called reproduction and only a few individuals, with highest fitness values are chosen to act as parents in-order to ensure that only the fittest genes are passed into the next generation.
These steps are repeated multiple times until some completion criteria is met. 

Mathematically EAs can be viewed as a class of black-box stochastic optimization techniques~\cite{salimans2017evolution, deepnevol}.
The reason they are ``black-box" is because they do not make any assumptions about the structure of the underlying function being
optimized, they can only evaluate it (like a lookup function).
This leads to the fundamental difference between RL and EA. Both try to optimize the expected reward, but RL perturbs the
action space and uses backpropagation (which is computation and memory heavy) to compute parameter updates,
while EA perturbs the parameter space (e.g., nodes and connections inside a NN) directly.
The ``black-box" property makes EAs highly robust - the same algorithm can learn how to solve various problems 
as from the algorithm's perspective the task in hand remains the same: perturb the parameters to maximize reward.

\subsection{The NEAT Algorithm}
\label{subsec:neat}

TWEANNS are a class of EAs which evolve both the topology and weights for given NN simultaneously.  
\textit{Neuro-Evolution for Augmented Topologies} (NEAT) is one of the algorithms in this class developed by Stanley et al \cite{neat}. 
We use NEAT to drive the system architecture of \system in this work, though it can be extended to work with other TWEANNs as well.
\autoref{fig:evol_algo}(b) depicts the steps and flow of the NEAT algorithm, and \autoref{fig:evol_algo}(d) lists the terminology we will use throughout 
this text.

{\bf Population.} The population in NEAT is the set of NN topologies in every generation that each run in the environment to collect a fitness score.

{\bf Genes.} The basic building block in NEAT is a {\it gene}, which can represent either a NN node (i.e., neuron), or a connection (i.e., synapse), as shown in \autoref{fig:evol_algo}(c).
Each node gene can uniquely be described by an id, the nature of activation (e.g., ReLU) and the bias associated with it.
Each connection can be described by its  starting and end nodes, and its hyper-parameters (such as weight, enable).

{\bf Genome}. A collection of genes that uniquely describes one NN in the population, as \autoref{fig:evol_algo}(c) highlights.

{\bf Initialization.} NEAT starts with a initial population of very simple topologies comprising only the input and the output layer.
It evolves into more complex and sophisticated topologies using the {\it mutation} and {\it crossover} functions.

{\bf Mutation.} Akin to biological operation, mutation is the operation in which a child gene is generated by tweaking the parameters of the parent gene.
For instance, a connection gene can be mutated by modifying the weight parameter of the parent gene.
Mutations can also involve addition or deletion of genes, with a certain probability.

{\bf Crossover.} Crossover is the name of the operation in which a child gene for the next generation is created by cherry picking 
parameters from two parent genes.

{\it Speciation and Fitness Sharing.} Evolutionary algorithms in essence work by pitting the individuals against each other in a given population and competitively selecting the fittest. 
However, it is not difficult to see that this scheme can prematurely prune individuals with useful topological features 
just because the new feature has not been optimized yet and hence did not contribute to the fitness. 
NEAT has two interesting features to counteract that, called {\it speciation} and {\it fitness sharing}. 
Speciation works by grouping a few individuals within the population with a particular niche. Within a species, the fitness of the younger individuals is artificially increased so that they are not obliterated when pitted against older, fitter individuals, thus ensuring that the new innovations are protected for some generations and given enough time to optimize.
Fitness sharing is augmenting fitness of young genomes to keep them competitive.

\vspace{-2mm}
\section{Computational Behavior of EAs}
\label{sec:motivation}
\vspace{-1mm}
This section characterizes the computational behavior of EAs, using NEAT as a case study, providing specific insights relevant for computer architects.

\begin{table*}[t]
\centering
\setlength{\abovecaptionskip}{0pt}
\setlength{\belowcaptionskip}{0pt}
\scriptsize
\caption{\small Open AI Gym~\cite{openai} environments for our experiments.}
\begin{tabular} {|p{1.5cm}|p{8cm}| p{3cm}| p{3cm}|}
\hline
\bf Environment & \bf Goal & \bf Observation & \bf Action\\
\hline
\textit{Acrobot}
& Balance a complex inverted pendulum constructed by linking two rigid rods 
& Six floating point numbers
& One floating point number
\\
\hline
\textit{Bipedal}
& Evolve control for locomotion of a two legged robot on a simple terrain. 
& Twenty four floating point numbers.
& Six floating point numbers.
\\ 
\hline
\textit{Cartpole\_v0} 
& The winning criteria is to balance an inverted pendulum on a moving platform for 100 consecutive time steps. 
& Four floating point numbers.
& One binary value.
\\
\hline
\textit{MountainCar}
& The goal of this task is to control an underpowered car sitting in a valley such that it reaches the finish point on the peak of one of the mountains.
& Two floating point numbers.
& One integer, less than three, for the direction of motion.
\\
\hline
\textit{LunarLander} 
& The goal to control the landing of a module to a specific spot on the lunar surface by controlling the fire sequence of its fours thrusters. 
& Eight floating point numbers.
& One integer, less than four, indicating the thruster to fire.
\\
\hline
\textit{Atari games} 
& The agent has to play Atari games by controlling button presses.
We used \textit{Airraid\_ram, Alien\_ram, Asterix\_ram and Amidar\_ram} environments 
& 128 bytes indicating the current state of the game RAM.
& One integer value, indicating the button press.\\ 
\hline
\end{tabular}
\label{table:environments}
\end{table*}

\insertWideFigure{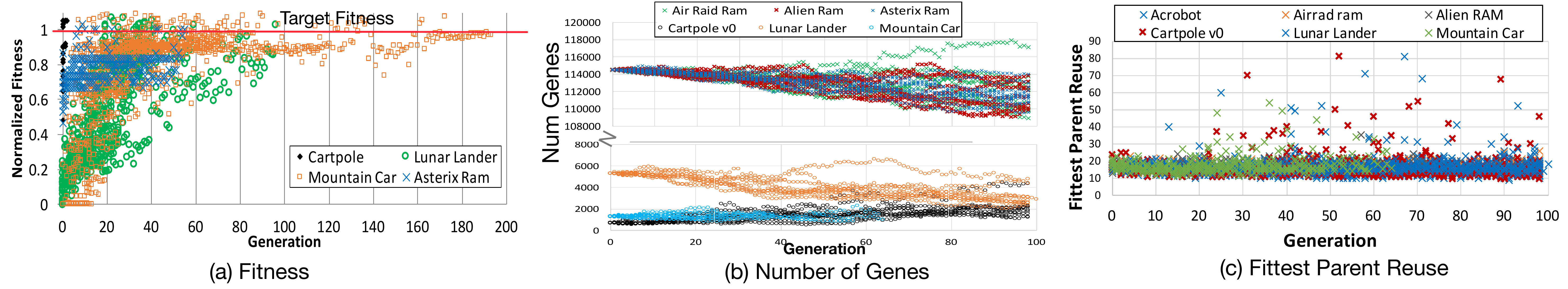}{Evolution behavior of Open AI Gym games as a function of generation.}

\insertWideFigure{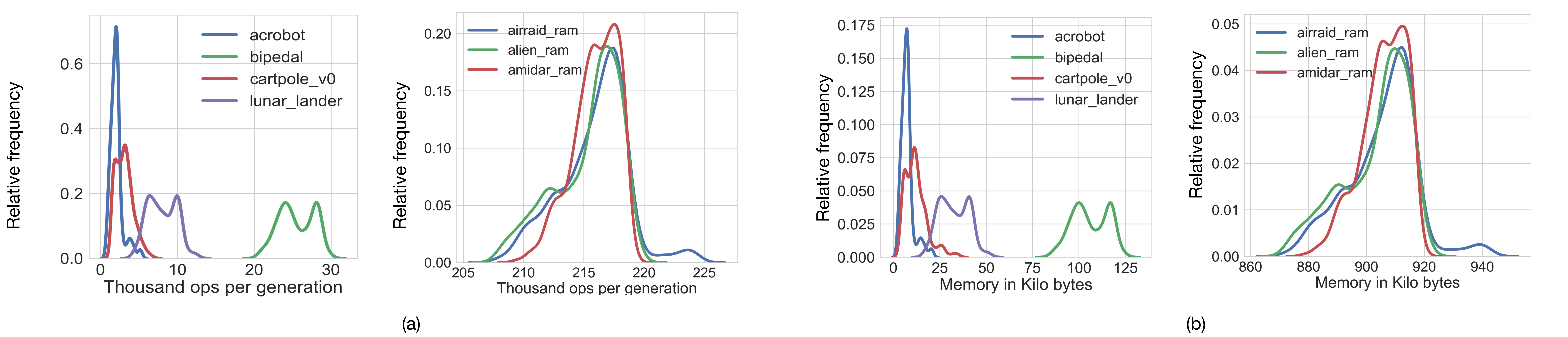}{(a) Computation (i.e., Crossover and Mutations) Ops and (b) Memory Footprint of applications from OpenAI Gym in every generation. A distribution is plotted across all generations till convergence and 100 separate runs of each application.}

\vspace{-1mm}
\subsection{Target Environments}
\label{sec:environments}
\vspace{-2mm}
We use a suite of environments described in \autoref{table:environments} from OpenAI gym~\cite{openai}. Each of these environments involves a learning task, which 
we ran through an open-source python implementation of NEAT~\cite{neat_python}.

\subsection{Accuracy and Robustness}
\label{sec:robustness}
\vspace{-2mm}
All experiments start with the same simple NN topology - a set of input nodes
(equal to the observation space of the environment) 
and a set of output nodes 
(equal to the action space of the environment).
These are fully-connected but the weight on each connection is set to zero.
We ran the same codebase for different applications, changing only the fitness function between these different runs.
All environments reached the target fitness - demonstrating the robustness of 
NEAT\footnote{We also ran the same environments with open-source implementations of A3C and DQN, two popular RL algorithms, and found that certain OpenAI environments never converged, or required 
a lot of tuning of the RL parameters for them to converge. However, a comprehensive comparison of RL vs. NE is beyond the scope of this paper.}.

\autoref{fig:plot_generation}(a) demonstrates the evolution behavior of four of these environments across multiple runs. 
We make two observations.
First, across environments, there can be variance in the average number of generations it takes to converge.
Second, even within the same environment, some runs take longer than others to converge, since the evolution process is probabilistic.
For e.g., for {\it Mountain Car}, the target fitness could be realized as early as generation \# 8 to as late as generation \# 160.
These observations point to the need for energy-efficient hardware to run NE algorithms as their total runtime can vary depending on the specific task they are trying to solve.

\subsection{Compute Behavior and Parallelism}
\label{sec:compute_trends}
\vspace{-2mm}
As shown in \autoref{fig:evol_algo}(a), EAs essentially comprise of an outer loop running the evolutionary learning algorithm to create new genomes (NNs) every generation,
 and inner loops performing the inference for these genomes.
 Prior work has shown that the computation demand of EAs drops by two-thirds compared to backpropagation~\cite{salimans2017evolution}.

\subsubsection{Learning (Evolution)}
\label{sec:compute_evolution}
\vspace{-2mm}
In NEAT, there are primarily two classes of computations that occur - crossover and mutation, as shown in \autoref{fig:evol_algo}(b).
\autoref{fig:plot_compute_memory}(a) show the distribution of the number of crossover and mutations operations within a generation.
The distribution is plotted across all generations till the application converged and across 100 runs of each application. 
We observe that the mutations and crossovers are in thousands in one class of applications, and are 
in the range of hundred thousands in another class. 
A key insight is that {\it crossover and mutations of each gene can occur in parallel}. 
This demonstrates a class of raw parallelism provided by EAs that prior work on 
accelerating EAs~\cite{salimans2017evolution, deepnevol} 
has not leveraged.
We term this as {\bf gene level parallelism (GLP)} in this work.
Moreover, as the environments become more complex with larger NNs with more genes, the amount of GLP actually increases!

\subsubsection{Inference}
\label{sec:compute_inference}
\vspace{-1mm}
The inference step of NEAT involves running inference through all NNs in the population, for the particular environment at hand.
Inference in NEAT however is different than that in traditional Multilevel Level Perceptron (MLP) NNs. Recall that NEAT starts with a simple topology (\autoref{sec:robustness}) and then adds new connection and nodes via mutation. This way of growing the network results in a irregular topology; or when viewed from the lens of DNN inference - a highly sparse topology.
Inference on such topologies is basically processing an acyclic directed graph.
An interesting point to note is that, following an evolution step, multiple genomes undergo the inference step concurrently (\autoref{fig:evol_algo}(a)). As there is no dependence within the genomes, a different opportunity of parallelism arises.
We term this as {\bf population level parallelism (PLP)}.  

\subsection{Memory Behavior}
\label{sec:memory_trends}
\vspace{-1mm}
\autoref{fig:plot_generation}(b)  plots the total number of genes as the NN evolves.

\subsubsection{Memory Footprint.}
\label{sec:memory_footprint}
\vspace{-1mm}
It is important to note that the memory footprint for EAs at any time is simply the space required to store all the genes of all genomes within a generation.
The algorithm does not need to store any state from the previous generations (which effectively gets passed on in the form of children) to perform the learning.
From a learning/training point of view, this makes EAs highly attractive - they can have much lower memory footprint than BP,
which requires error gradients and datasets from past epochs to be stored in order to run stochastic gradient descent.
From an inference point of view, however, the lack of regularity and layer structure means that genomes cannot be encoded as efficiently 
as convolutional neural networks today are.
There have been other NE algorithms such as HyperNEAT~\cite{hyperneat} which provide a mechanism to encode the genomes more efficiently, 
which can be leveraged if need be.

For all the applications in the Open AI gym we looked at, the overall memory footprint per generation was less than 1MB, as \autoref{fig:plot_compute_memory}(b) shows.
While larger applications may have larger memory footprints per generation, 
the total memory is still expected to be much less than that required by 
training algorithms due to the reasons mentioned above, enabling a lot of the memory required by the EA to be cached on-chip.

\subsubsection{Communication Bandwidth}
\label{sec:comm_bw}
\vspace{-1mm}
Leveraging GLP and PLP requires streaming millions of genes to compute units, increasing the memory bandwidth pressure. 
Caching the necessary genes/genomes on-chip, and leveraging a high-bandwidth network-on-chip (NoC) can help provide this bandwidth, as we demonstrate via \system.

\subsubsection{Opportunity for Data Reuse}
\label{sec:glr}
\vspace{-2mm}
Data reuse is one of the key techniques used by most accelerators~\cite{chen2016eyeriss}.
Unlike DNN inference accelerators which have regular layers like convolutions that directly expose reuse across filter weights, 
the NN itself is expected to be highly  irregular in an evolutionary algorithm.
However, we identify a different kind of reuse: {\bf genome level reuse (GLR)}.
In every generation, the same fit parent is often used to generate multiple children.
We quantify this opportunity in \autoref{fig:plot_compute_memory}(c).
For most applications, the fittest parent in every generation was reused close to 20 times, and for some applications like Cartpole and Lunar lander, 
this number increased up to 80. In other words, one parent genome was used to generate 80 of the 150 children required in the next generation, 
offering a tremendous opportunity to read this genome only once from memory and store it locally.
This can save both energy and memory bandwidth.

\insertWideFigure{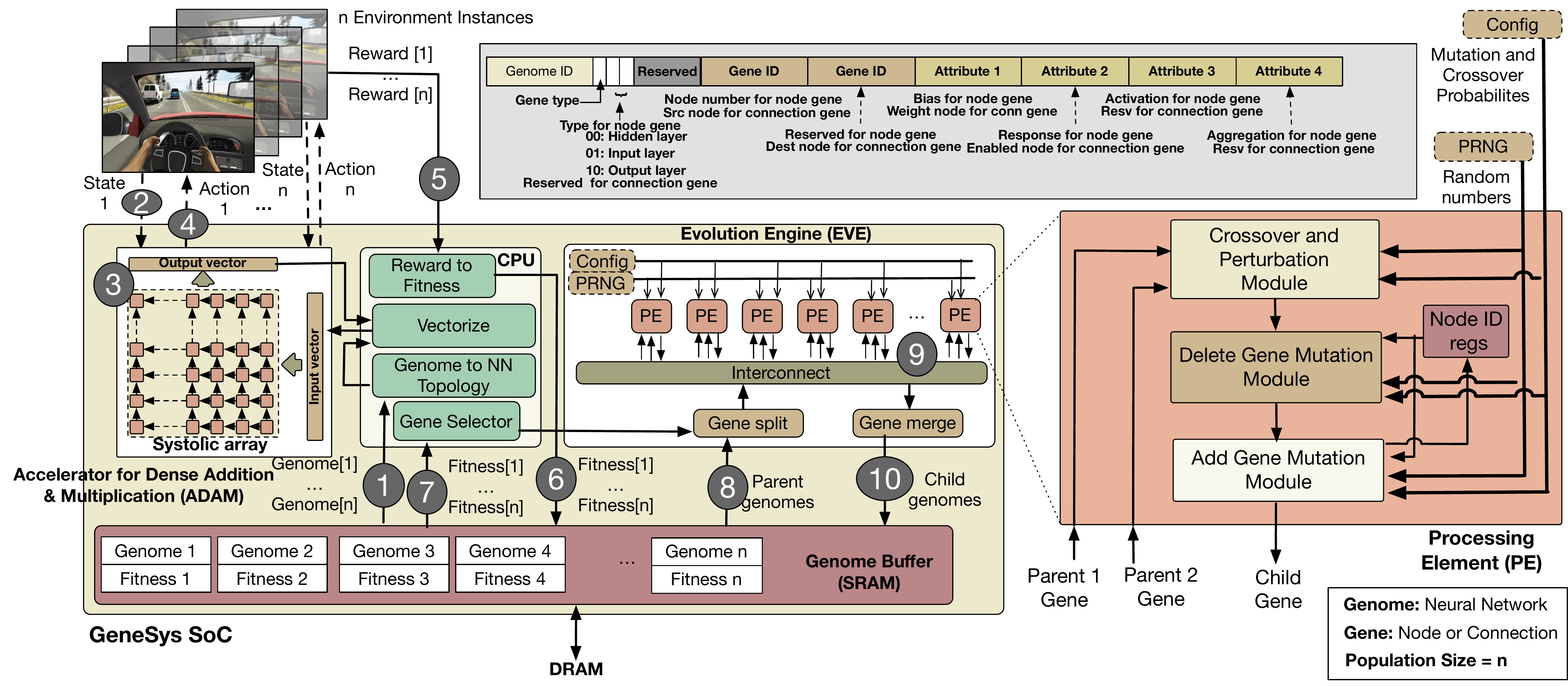}{Overview of the \system system: CPU for algorithm config, Inference Engine (ADAM), Learning Engine (EvE) and EvE PE.}

\begin{table}[t]
\centering
\setlength{\abovecaptionskip}{0pt}
\setlength{\belowcaptionskip}{0pt}
\scriptsize
\caption{\small Comparing DQN with EA}
\begin{tabular} {|p{0.9cm}|p{3.7cm}| p{3.5cm}|}
\hline
\bf  & \bf DQN & \bf EA\\
\hline
\textit{Compute}
& 3M MAC ops in forward pass, 680K gradient calculations in BP
& 115K MAC ops in inference, 135K crossover + mutations in evolution\\ 
\hline
\textit{Memory}
& 50 MB for replay memory of 100 entries, 4 MB for parameters and activation given mini-batch size of 32
& $\textless$1MB to fit entire generation\\ 
\hline
\textit{Parallelism}
& MAC and gradient updates can parallelized per layer
& GLP and PLP as described in ~\autoref{sec:compute_inference} and ~\autoref{sec:compute_evolution}\\ 
\hline
\textit{Regularity}
& Dense CNN with high regularity and opportunity of reuse
& Highly sparse and irregular networks\\ 
\hline
\end{tabular}
\label{table:comparisons}
\end{table}

\subsection{A case for acceleration}
\vspace{-2mm}
In this section we present the key takeaways from the compute and memory analysis of EA. We also compare compute-memory requirements of EA with conventional RL in ~\autoref{table:comparisons} with DQN~\cite{dqn} as a candidate, both running ATARI. 

We notice that EA has both low memory and compute cost when compared to DQN. 
Given the the reasonable memory foot print  (less than 1MB for the applications we looked at) and GLR opportunity, it is evident that a sufficiently sized on chip memory can help remove/reduce off-chip accesses significantly, saving both energy and bandwidth.
Also the compute operations in EA (crossover and mutations) are simple and hardware friendly. Furthermore, the absence of gradient calculation and significant communication overheads facilitate scalability ~\cite{salimans2017evolution, deepnevol}.
The inference phase of EAs is akin to graph processing or sparse matrix multiplication, and not traditional dense GEMMs like conventional DNNs, dictating the choice of the hardware platform on which they should be run.

If we can reduce the energy consumption of the compute ops by implementing them in hardware, pack a lot of compute engines in a small form factor, 
and store all the genomes on-chip, complex behaviors can be evolved even 
in mobile autonomous agents. This is what we seek to do with \system, which we present next.
\section{Genesys: System and Microarchitecture}
\label{sec:genesys}


\subsection{System overview}
\label{subsec:system}
\vspace{-1mm}

\system is a SoC for running evolutionary algorithms in hardware.
This is the first system, to the best of our knowledge, to perform evolutionary learning and inference on the same chip.
\autoref{fig:Genesys_soc_pe} present an overview of our design.
There are four main components on the SoC:
\squishlist
\item{\textit{Learning Engine (\textbf{EvE})}}: EvE is the accelerator proposed in this work. It is responsible for carrying 
out the selection and reproduction part of the NEAT algorithm parts of the NEAT algorithm across all genomes of the population. It consists of a collection of processing elements (PEs), designed for power efficient implementation of crossover and mutation operations. Along with the PEs, there is a gene split unit to split the parent genome into individual genes, an on-chip interconnect to send parent genes to the PEs and collect child genes, and a gene merge unit to merge the child genes into a full genome. 
\item{\textit{Inference Engine (\textbf{ADAM})}}: 
We observed in \autoref{sec:compute_inference},
 the neural nets generated by NEAT are highly irregular in nature. This irregularity deems traditional DNN accelerators unfit for inference in this case, as they are optimized with the assumption that the topology is a dense cascades of layers. 
In our case inference is closer to graph processing than DNN inference, which is essentially a sequence of multiple vertex updates for the nodes in the NN graph. 

\adam consists of a systolic array of MAC units to perform parallel vertex evaluations, and a vectorize routine in System CPU to pack nodes into well formed input vectors for dense matrix-vector multiplication.
Similar to input vector creation, the vectorize routine also generates weight matrices for genomes, every time a new generation is spawned. 
However, as the weight matrices do not change within a given generation, and are reused for multiple inferences, while every new vertex evaluation requires a new input vector. 

\item{\textit{System CPU (ARM Cortex M0 CPU)}}: We use an embedded Cortex M0 CPU to perform the configuration steps of the 
NEAT algorithm (setting the various probabilities, population size, fitness equation, and so on), and manage data conversion and movement between EvE, ADAM and the on-chip SRAM. 
\item{\textit{Genome Buffer (SRAM)}}: We use a shared multi-banked SRAM that harbors all the genomes for a given generation and is accessed by both ADAM and EvE. This is backed by DRAM for cases when the genomes do not fit on-chip.
\squishend

\subsection{Walkthrough Example}
\vspace{-2mm}
\label{subsec:walkthrough}
We present a brief walk-through of the execution sequence in the system with the help of \autoref{fig:Genesys_soc_pe} to demonstrate the dataflow through the system.
Our system starts with a population of genomes of generation \textit{n} in memory. Through the set of steps described, next,  \system  evolves the genomes for the next generation \textit{n + 1}.
\squishlist
\item{\textbf{Step 1:} The genomes (i.e., NNs) are read from the genome buffer SRAM and mapped over the MAC units in ADAM.}
\item{\textbf{Step 2:} ADAM reads the state of the environment. In our evaluations, the environment is one of the OpenAI gym games (\autoref{table:environments}).}
\item{\textbf{Step 3:} Inference is performed by multiple vertex update operations. Several vertices are simultaneously updated by packing input vertices into a well formed vector in the CPU, followed by matrix-vector multiplication on systolic array. Inference for a given genome is marked as complete once the output vertices are updated.}
\item{\textbf{Step 4:} The output activations from \textit{step 3} are translated as actions and fed back to the environment.}
\item{\textbf{Step 5:} Steps 2-4 are repeated multiple times until a completion criteria is met. For the OpenAI runs, this was either a success or failure in the task at hand. Following this a cumulative reward value is obtained from the environment - a proxy for performance of the NN.}
\item{\textbf{Step 6:} The reward value is then translated into a fitness value by the CPU thread. The reward depends upon the application/environment. The fitness value is augmented to the genome that was just run in SRAM.}
\item{\textbf{Step 7:} Once the fitness values for all individuals in the population are obtained, reproduction for the next generation can now start. In NEAT, only individuals above a certain fitness threshold area are allowed to participate in reproduction. A selector logic running on the CPU takes these factors into account and selects the individuals to act as parents in the next generation.}
\item{\textbf{Step 8:} The selected parent genomes are read by EvE. The gene splitting logic curates genes from different parents that will produce the child genome, aligns them, and stream them to the PEs in EvE.}
\item{\textbf{Step 9:} The PEs receive the parent genes from the interconnect, perform crossovers and mutations to produce the child genes, and send these genes back to interconnect.}
\item{\textbf{Step 10:} The gene merge logic organizes the child genes and produces the entire genome. Then this genome is written back into the genome buffer, overwriting the genomes from the previous generation. As each child genome becomes ready, it can be launched over ADAM once again, repeating the whole process.}
\squishend

The system stops when the CPU detects that the target fitness for that application has been achieved.
Steps 1 to 6 can leverage PLP, while steps 8 to 10 can leverage GLP. Step 7 (fittest parent selection) is the only serial step.

\subsection{Microarchitecture of EVE}
\label{subsec:microarch}

\subsubsection{Gene Level Parallelism (GLP)}
\vspace{-2mm}

We leverage parallelism within the evolutionary part - namely at the gene level.
As discussed earlier, the operations in an EA can broadly be categorized in two classes: crossover and mutation.
In NEAT, there are three kinds of mutations (perturbations, additions and deletions).
These four operations are described in \autoref{fig:evol_algo}(d).
While these four operations themselves are serial, they do not have any dependence with other genes.
Moreover, the high operation counts per generation (\autoref{fig:plot_compute_memory}(a)) 
indicates massive GLP which we exploit in our proposed microarchitecture via multiple PEs.

\vspace{-1mm}
\subsubsection{Gene Encoding}
\label{subsubsec:ds}
\vspace{-2mm}

\autoref{fig:Genesys_soc_pe} shows the structure for a gene we use in our design.
NEAT uses two types of genes to construct a genome, 
a node gene which describe vertices and the connection gene which describe the edges in the neural network graph. 
We use 64 bits to capture both types of genes. 
Node genes have four attributes - \{Bias, Response, Activation, Aggregation\}~\cite{neat}.
Connection genes have two attributes - source and destination node ids.
%

\vspace{-1mm}
\subsubsection{Processing Element (PE)}
\label{subsubsec:pe}
\vspace{-2mm}
\insertWideFigure{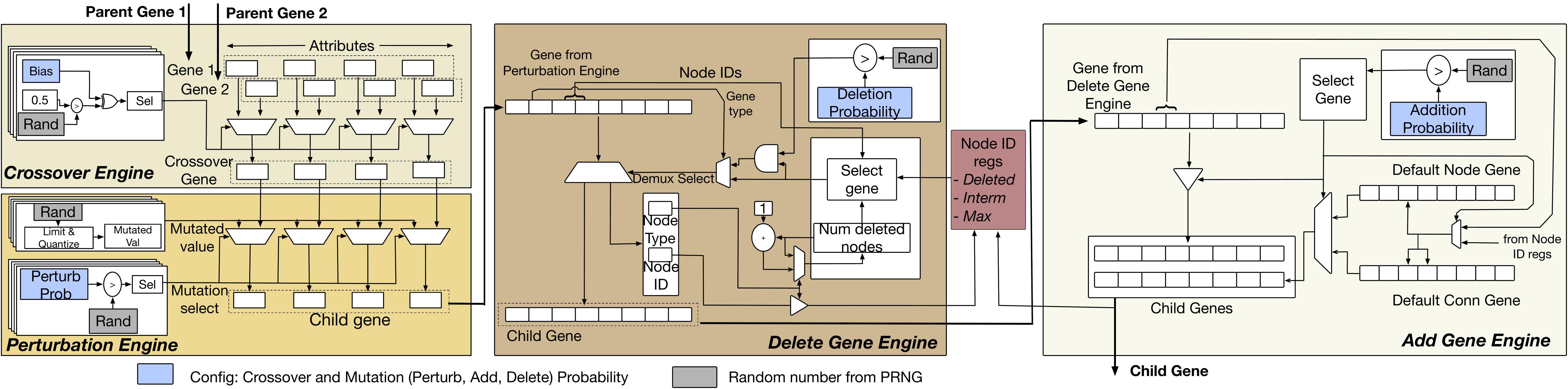}{Schematic depicting the various modules of the Eve PE. }



\autoref{fig:Genesys_soc_pe} shows the schematic of the EvE PE. It has a four-stage pipeline. These stages are shown in 
\autoref{fig:pe_blocks}.
\textit{Perturbation, Delete Gene} and \textit{Add Gene} are three kinds of mutations that our design supports.

\textbf{Crossover Engine.}  The crossover engine receives two genes, one from each parent genome.
As  described in \autoref{subsec:neat}, crossover requires picking different attributes from the parent genome to construct the child genome. 
The random number from the PRNG is compared against a {\it bias} and used to select one of the parents for each of the attributes.
We provide the ability to program the bias, depending on which of the two parents contributes more attributes (i.e., is preffered) 
to the child. The default is 0.5.
This logic is replicated for each of the 4 attributes. 

\textbf{Perturbation Engine}.
A perturbation probability is used to generate a set of mutated values for each of the attributes in the child gene that was generated by the crossover engine.

\textbf{Delete Gene Engine}. 
There are two types of genes in a given genome - node and connection -
and  implementing gene deletion for each of them differs. 
Irrespective of the type, the decision to delete a gene is taken by comparing the deletion probability with a number generated by PRNG.
%
For node deletion, in addition to the probability, the number of previously deleted nodes is also checked. 
If a threshold amount of nodes are previously deleted, no mode deletion happens in order to keep the genome alive.
If not then the node is nullified and its ID is stored. This ID is later compared with the source and destination IDs of any of the connection genes to ensure no dangling connection exist in the genome.
Deletion of connections, is fairly straight forward, but deletion decision is taken either by comparing the gene IDs as mentioned above or by comparing deletion probabilities.

\textbf{Add Gene Engine}.
This is the fourth and final stage of the PE pipeline. 
As in the case of the previous stage, depending upon the type of the gene, the implementation varies.
To add a new node gene, the logic inserts a new gene with default attributes and a node ID greater than any other node 
present in the network. Additionally two new connection genes are generated and the incoming connection gene is dropped.
The addition of a new connection gene is carried out in two cycles. When a new connection gene arrives, the selection logic compares a random number with the addition probability. If the random number is higher, 
then the source of the incoming gene is stored. 
When the next connection gene arrives, the logic reads the destination for that gene, appends the stored source value and default attributes, and creates a new connection gene. 
This mechanism ensures that any new connection gene that is added by this stage 
always has valid source and destinations.

\vspace{-1mm}
\subsubsection{Gene Movement}
\vspace{-2mm}
Here, we describe the blocks that manage gene movement.

\textbf{Gene Selector}.
As we discussed in \autoref{subsec:neat}, only a few individuals in a given population get the opportunity to contribute towards the reproduction of the next generation. In very simple terms, selection is performed by determining a fitness threshold and then eliminating the individuals below the threshold. 
In \autoref{subsec:neat} we have seen that NEAT provides a mechanism to keep new features in the population by speciation and fitness sharing. 
The selection logic in our design works in three steps. First, the fitness values of the individuals in the present generation and read and adjusted to implement fitness sharing. 
Next, the threshold is calculated using the adjusted fitness values. 
Finally the parents for the next generation are chosen and the list of parents for the children is forwarded to the gene splitting logic.
This is handled by a software thread on the CPU, as shown in \autoref{fig:Genesys_soc_pe}.

\textbf{Gene Split.}.
The Gene Split block orchestrates the movement of genes from the Genome Buffer to the PEs inside EvE.
In the crossover stage, the keys (i.e., node id) for both the parent genes need to be the same. 
However both the parents need not have the same set of genes or there might be a misalignment between the genes with the same key among the participating parents. 
The gene split block therefore sits between the PEs and the Genome Buffer to ensure that the alignment is maintained and proper gene pairs are sent to the PEs every cycle. 
 
In addition, this block receives the list of children and their parents from the Gene Selector and takes care of assigning the PEs to generate the child genome.
We describe the assignment policy and benefits in \autoref{subsec:impl}.

\textbf{Gene Merge}.
Once a child gene is generated, it is written back to the Gene Memory as part of the larger genome it is part of. This is handled by the Gene Merge block.

\textbf{Pseudo Random Number Generators (PRNG)}.
The PRNG feeds  a 8-bit random numbers every cycle to all the PEs, as shown in \autoref{fig:Genesys_soc_pe}.
We use the XOR-WOW
 algorithm, also used within NVIDIA GPUs, to implement our PRNG.

\textbf{Network-on-Chip (NoC)}
A NoC manages the distribution of parent genes from the Gene Split to the PEs and collection of child genes at the Gene Merge.
We explored two design options for this network.
Our base design is separate high-bandwidth buses, one for the distribution and one for the collection
However, recall that the NEAT algorithm offers opportunity for reuse of parent genomes across multiple children, as we showed in \autoref{sec:glr}.
Thus we also consider a tree-based network with multicast support
and evaluate the savings in SRAM reads in \autoref{sec:eval}.


\subsubsection{Integration}
\label{subsec:impl}
\vspace{-2mm}
In this section we will briefly describe how the different components are tied together to build the complete system. 

\textbf{Genome organization.} As described in earlier sections, we have two types of genes, nodes and connection. 
As shown in \autoref{fig:Genesys_soc_pe} each gene can be uniquely identified by the gene IDs. 
In this implementation we identify node genes with positive integers, and the connection genes by a pair of node IDs representing the source and the destination. 
Within a genome, the genes are stored in two logical clusters, one for each type.
Within each cluster, the genes are stored by sorting them in ascending order of IDs. 
Ensuring this organization eases up the implementation of the Add Gene engine. 
During reproduction, since the child gene gets the key of the parent genes, which in turn are streamed in order, ordering is maintained.
For newly added genes, the Gene Merge logic ensures that they sequenced in the right order when put together in memory.

\textbf{EvE Dataflow.} After the Gene Selector finalizes the parents and their respective children, 
the list is passed to the Gene Split block. The Gene Split logic then allocates PEs for generation of the children. 
In this implementation we allocate only one PE per child genome\footnote{It is possible to spread the genome across multiple PEs as well but might lead to different genes 
of a genome arriving out-of-order at the Gene Merge block complicating its implementation.}.
The PE allocation is done with a greedy policy, such that maximum number of children can be created from the parents currently in the SRAM.
This is done to exploit the reuse opportunity provided by the reproduction algorithm and minimize SRAM reads. 

When streaming into the PE, the node genes are streamed first. 
This is done in order to keep track of the valid node IDs in the genome, which will then be used in the gene addition and deletion mutations. 
Information about valid nodes are required to prune out dangling connections and assignment of node IDs in case of a new node or connection addition.
Once the nodes are streamed, connection genes are streamed until the complete genome of the child is created.
Before the genes are streamed, it takes 2 cycles to load the parents' fitness values and other control information.

\insertWideFigure{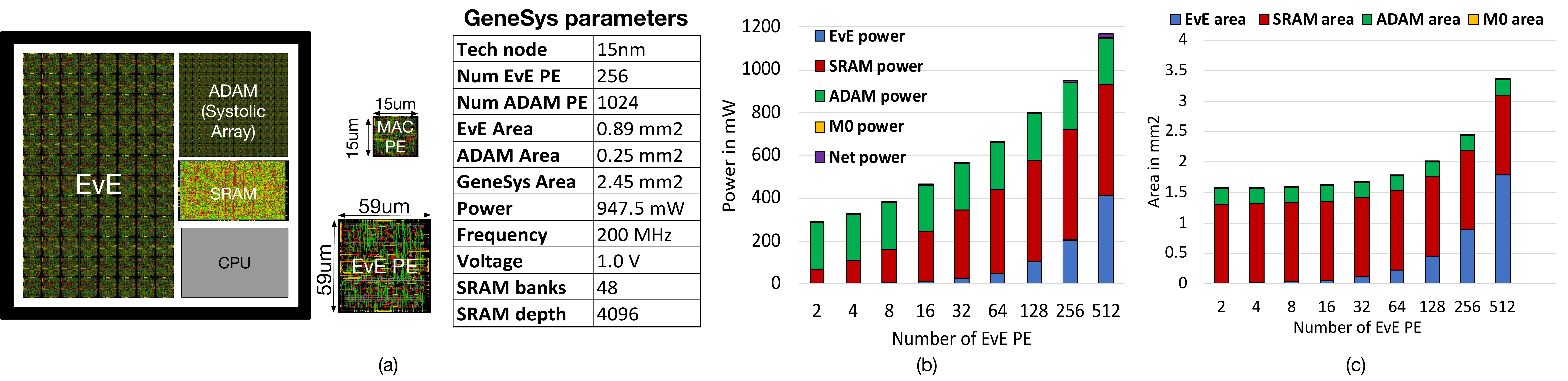}{(a) Place-and-Routed GeneSys SoC (b) Power consumption with increase in PE in EvE (c) Area footprint with increase in PE in EvE.}

\subsection{Microarchitecture of ADAM}
\label{subsec:microarch_adam}
\vspace{-2mm}
As mentioned in \autoref{subsec:system}, \adam evaluates NNs generated by \eve by processing vertices in the irregular NN graph.
We had two design choices - either go with a conventional graph accelerator like \textit{Graphicionado} \cite{ham2016graphicionado}, 
or pack the irregular NN into 
dense matrix-vector multiplications.
Recall that EAs have a small memory requirement (unlike conventional graph workloads) and do not require caching optimizations.
Moreover, given that our workloads are neural networks, 
vertex operations are nothing but multiply and accumulate.
We thus decided to go with the latter approach.
\adam performs multiple vertex updates concurrently, 
by posing the individual vector-vector multiplications into a packed matrix-vector multiplication problem.
Systolic array of Multiply and Accumulate (MAC) elements is a well known structure for energy efficient matrix-vector multiplication in hardware, and is essentially the heart of \adam's microarchitecture.

However, picking the ready node values to create input vectors for packed matrix-vector multiplication is a task with heavy serialization.
We use the System CPU to generate required vectors from the node genomes. 
As both systolic arrays and graph processing are heavily investigated techniques in literature~\cite{kung1980algorithms, moldovan1983design, ham2016graphicionado, ahn2016scalable, dai2016fpgp, han2013turbograph}, we omit details of implementation for the sake of brevity.

\vspace{-2mm}
\section{Implementation}
\label{sec:eve_impl}
\vspace{-2mm}


{\bf \system SoC. }We implemented the \system SoC using Nangate 15nm FreePDK.
We implement a $32\times32$ systolic-array of MAC units for ADAM and measure the post synthesis power and area numbers. 
EvE PEs are synthesized and the area and power numbers are recorded similar to ADAM, as shown in \autoref{fig:Scalability_Plots}(a). \autoref{fig:Scalability_Plots}(b) shows the roofline power as function of EvE PEs. We call it roofline because the numbers here are calculated on the assumption that \system is always computing and thus capture the maximum; actual power will be much lower. In later sections we will discuss why this an overly pessimistic assumption and ways power consumption can be lowered.
Motivated by the memory footprint in \autoref{sec:memory_footprint}, we allocated 1.5MB for on-chip SRAM. The SRAM has 48 banks to exploit the reuse of parents observed in \autoref{sec:glr}, as well as to reduce conflict while feeding data to ADAM.
We also take into account the area and power contributed by the interconnect and the cortex M0 processor core.
With the post synthesis numbers and the relationship of SRAM size and number of PEs, we generate the area footprint for design points with varying number of PE, \autoref{fig:Scalability_Plots}(c) depicts the numbers.

We choose the operation frequency to be 200MHz, which is typical of the published neural network accelerators ~\cite{isscc_2016_chen_eyeriss, sim201614, desoli201714, moons2017envision}.
With 256 PEs, we comfortably blanket under 1W as shown in \autoref{fig:Scalability_Plots}(b).

\vspace{-2mm}
\begin{table}[h]
	\centering
	\setlength{\abovecaptionskip}{-4pt}
\setlength{\belowcaptionskip}{-1pt}
	\vspace{-2mm}
	\caption{Target System Configurations.}
	\begin{tabular}{llll}
	\hline
		Legend & Inference & Evolution & Platform\\
	\hline
		CPU\_a & Serial & Serial & 6th gen i7\\
		CPU\_b & PLP & Serial & 6th gen i7\\
		GPU\_a & BSP & PLP & Nvidia GTX 1080\\
		GPU\_b & BSP + PLP & PLP & Nvidia GTX 1080\\
        		CPU\_c & Serial & Serial & ARM Cortex A57\\
		CPU\_d & PLP & Serial & ARM Cortex A57\\
		GPU\_c & BSP & PLP & Nvidia  Tegra\\
		GPU\_d & BSP + PLP & PLP & Nvidia Tegra\\
        \system & PLP & PLP + GLP & \system\\
	\hline
	\end{tabular}
\footnotesize{
PLP (GLP) - Population (Gene) Level Parallelism \\
BSP - Bulk Synchronous Parallelism (GPU)
}
\end{table}

\vspace{-2mm}
\section{Evaluation}
\label{sec:eval}
\vspace{-3mm}
  
%

\insertWideFigure{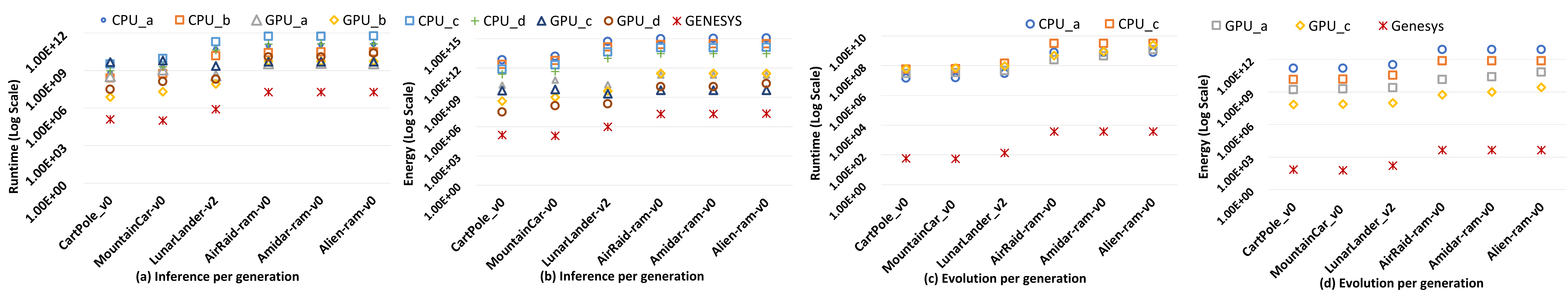}{Runtime and Energy for OpenAI gym environments across CPU, GPU and GeneSys. (a) Runtime, (b) Energy for Inference; and (c) Runtime, (d) Energy of Evolution}
\insertWideFigure{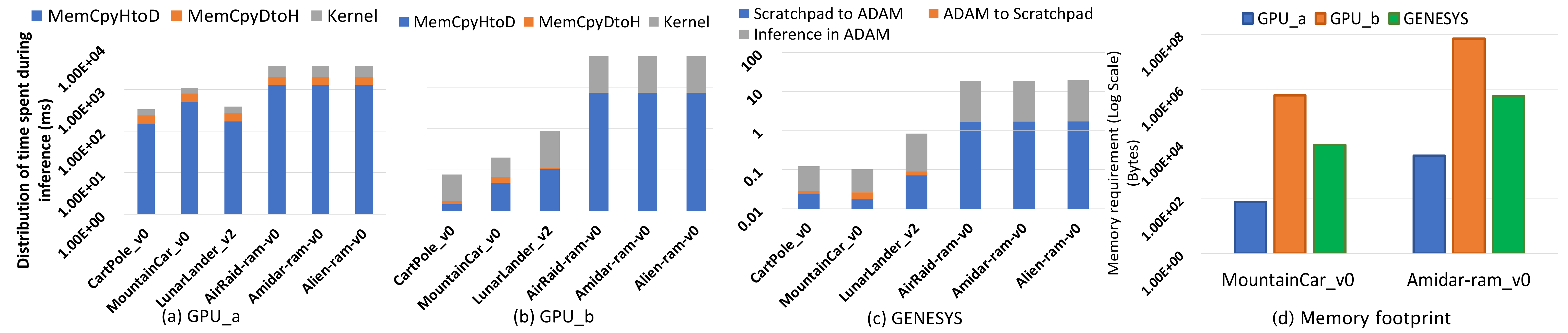}{Distribution of time spent in data-transfer and compute in (a) GPU\_a config, (b) GPU\_b config and (c) GENESYS; (d) depicts the variation in memory footprints for given application on various platforms}


\subsection{Methodology}
\label{subsec:method}
\vspace{-2mm}
We study the energy, runtime and memory footprint metrics for \system and compare these with the corresponding metrics in embedded and desktop class CPU and GPU platform.
For our study we use NEAT python code base ~\cite{neat_python}, and modify the evolution and inference modules as per our needs. 
We modify the code to optimize for runtime and energy efficiency on GPU and CPU platforms by exploiting parallelism and to generate a trace of reproduction operations for the various workloads presented in \autoref{table:environments}.


\textbf{CPU evaluations}. We measure the completion time and power measurements on two classes of CPU, desktop and embedded. 
The desktop CPU is a 6th generation intel i7, while the embedded CPU is the ARM Cortex A57 housed on Jetson TX2 board. 
On desktop, power measurements are performed using Intel's power gadget tool while on the Jetson board we use the onboard instrumentation amplifier INA3221.
We capture the average runtime for evolution and inference from the codebase, and use it to calculate energy consumption.

\textbf{GPU evaluations.} 
Similar to CPU measurements, we use desktop (nVidia GTX 1080) and embedded (nVidia Tegra on Jetson TX2) GPU nodes.
For the desktop GPU, power is measured using nvidia-smi utility while same onboard INA3221 is used for measuring gpu rail power on TX2.
Runtime is captured using nvprof utility for kernels and data-transfers, and are used in energy calculations.
To ensure that the correctness of the operations are maintained, we apply some constraints in ordering, for example crossovers precede mutation in time.
  

\textbf{\system evaluations.} The traces along with the parameters obtained by our analysis in \autoref{sec:eve_impl} are used to estimate the energy consumption for our chosen design point of EvE. 
Each line on the trace captures the generation, the child gene and genome id, the type of operation - mutation or crossover, and the parameters changed or added or deleted by the operations.
These traces serve as proxy for our workloads when we evaluate \eve and \adam implementations.


\vspace{-2mm}
\subsection{Runtime}
\vspace{-2mm}
\autoref{fig:performance_energy}(a) and (c) shows the runtime of different OpenAI gym environments on various platforms for both evolution and inference. 
In CPU, evolution happens sequentially while we try to exploit PLP in inference by using multithreading, running 4 concurrent threads (CPU\_b and CPU\_d). Parallel inference on CPU is 3.5 times faster than the serial counterpart.

We try to exploit maximum parallelism in GPU by mapping PLP and GLP to BSP paradigm in inference in two different implementations. Genesys outperforms the best GPU implementation by 100x in inference. 
Next, we describe our GPU implementations and discuss our observations.

\textbf{GPU deep dive.} GPU\_a exploits GLP by forming compaction on input vectors serially and evaluating multiple vertices in parallel for each genome. In GPU\_b, multiple vertices across genomes are evaluated in parallel thus exploiting both GLP and PLP. However the inputs and weights could no longer be compacted resulting in large sparse tensors.
\autoref{fig:GPU_vs_GENESYS}(a,b,c) depict the contribution of memory transfer in total runtime. We observed memory transfers take 70\% of runtime in GPU\_a, while GPU\_b takes to 20\% of total runtime for memory transfer.
\system in comparison also take about 15\% for memory transfers; however since all the data is on chip, the actual runtime is 1000x smaller. 
\autoref{fig:GPU_vs_GENESYS}(d) depicts the overall on-chip memory requirement in the GPU\_a, GPU\_b and \system. We see that GPU\_b has a much higher footprint as all sparse weight and input matrices are kept around, while for GPU\_a only compact matrices for one genome is required at a time. \system stores entire population in memory, thus we see 100x more footprint than GPU\_a, which is expected as we have a population size of 150. \system has 100x less footprint than both GPU\_b as GPU\_b as genomes rather than sparse-matrices are stored on chip.
\autoref{fig:eval_plots_param_sweep}(a) shows the distribution of connections and nodes in various workloads. The more the number of connection genes means denser weight matrices during inference hence higher utilization in ADAM. 

\subsection{Energy consumption}
\vspace{-1mm}
\autoref{fig:performance_energy}(b) and (d) shows the energy consumption per generations for OpenAI gym workloads on different platforms. 
ADAM contributes to 100x more energy efficiency, while EVE turns out to be 4 to 5 orders of magnitude more efficient than GPU\_c, the most energy efficient among our platforms.

\vspace{-1mm}
\subsection{Design choices: PEs, SRAMs and Interconnect}
\vspace{-1mm}
\insertWideFigure{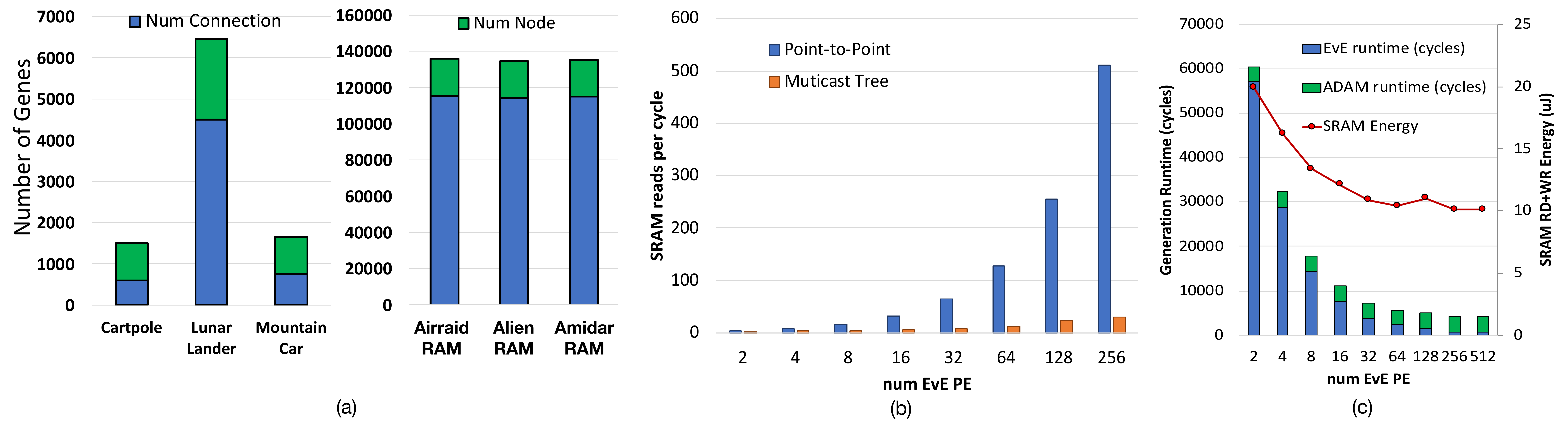}{(a) Composition of gene-types in genomes for different workloads (b) SRAM reads per cycle in Point-to-Point vs Multicast tree (c) SRAM energy consumption and runtime per generation as a function of number of EvE PE averaged for Atari workloads}
{\bf Impact of Network-on-Chip}
Neural network accelerators often take advantage of the reuse in data flow to reduce SRAM reads and hence lower the energy consumption. 
The idea is that, if same data is used in multiple PEs, there is a natural win by reading the data once and multicasting to the consumers. 
In our case, we see reuse in the parents while producing multiple children of a single parent. 
Therefore we can use similar methods to reduce reads as well. 
\autoref{fig:eval_plots_param_sweep}(b) shows the number of SRAM reads with a simple point-to-point network versus a multicast tree network.
We observe more than a 100$\times$ reduction in SRAM reads when supporting multicasts in the network, 
motivating an intelligent interconnect design.
An intelligent interconnect can also help support multiple mapping strategies of genes across the PEs, and is an interesting topic for future research.

{\bf Parallelizing Evolution}
Till now we have talked about EvE PE in terms of \textit{GLP} and reducing compute cost by implementing GA operations in hardware. This line of reasoning can lead to the question that weather \textit{GLP} can be traded-off for energy-benefits. The answer to this lies in \autoref{fig:eval_plots_param_sweep}(c), where we show the SRAM energy consumption for evolution (Read+Write) and generation time as a function of EvE PEs; size of ADAM and SRAM are constant. 
The SRAM energy curve indicates that there is almost monotonic improvement in energy efficiency as more EvE PEs are added. 
The linear decrease in energy (the curve shows exponential decrease for exponential increase in number of PEs) is a direct consequence of \textit{GLR}.
At lower PE counts, child genomes sharing same parent PEs are generated over time, thus requiring a single operand to be read over and over again. As the number of PEs increase multiple children sharing the same parent can be serviced by one read if we employ an appropriate interconnect capable of multicasting. 

Diverting attention to the runtime plot reveals a couple of interesting trends.
First the cycle count for inference is far less than intuitively expected for typical neural networks.
This is attributed to two factors, (i) The networks generated by NEAT are significantly simple and small than traditional Deep MLPs, and (ii) ADAM's high throughput aids fast Vector-Matrix computations we use to implement vertex updates. 
The other more interesting trend that we see is that at lower EvE PE counts the evolution runtime is disproportionately larger than inference!
The exponential fall off depicts that performance wise evolution is compute-bound, which is in agreement to our observations on \textbf{GLP} and \textbf{PLP} in \autoref{sec:compute_trends}
 
Decreasing the generation runtime has further benefits than it meets the eye. In our work we used simulated environments with which we can interact instantly. However, for real life workloads, the interactions will be much slower. This enables us to use circuit level techniques like clock and power gating to save even more power. The lower the compute window for \system the more time is used to interact with the environment thus saving more energy as we hinted in \autoref{sec:eve_impl}.

The tapering off of the trends in \autoref{fig:eval_plots_param_sweep}(c) at 256 PEs is due to the fact that we exploit only PLP for our experiments and at population size of 150 we intentionally restrict the exploitable parallelism.

\vspace{-2mm}
\section{Discussion and Related Work}
\label{sec:related_work}
\vspace{-2mm}
{\bf Future Directions.} It is important to note that the success of evolutionary algorithms is tied to the nature of application.
From a very high level what EA does, is search for optimal parameters guided by the fitness function and reward value. 
Naturally, as the parameter space for a problem becomes large, the convergence time of EAs increase as well.
In such a scenario, we believe that  \system can be run in conjunction with supervised learning, 
with the former enabling rapid topology exploration and then  using conventional training to tune the weights.
Neuro-evolution to generate deep neural networks ~\cite{bayer2009evolving, verbancsics2013generative, genetic_cnn, real2017largescale, ghazi2017nmode, deepneat2017} falls in this category.
The only thing that would change is the definition of \textit{gene}.
  
 
\textbf{Neuro-evolution.} Research on EAs has been ongoing for several decades.
~\cite{taylor1997selecting, larkin2006towards, ding2010using, sher2010dxnn} are some examples of early works in using evolutionary techniques for topology generations. Apart from NEAT ~\cite{neat}, other algorithms like Hyper-NEAT and CPPN ~\cite{hyperneat, cppn} for evolution of NNs have also been reported in the last decade ~\cite{d2008generative, hypernetworks, dppn}.

{\bf Online Learning.}  Traditional reinforcement learning methods have also gained traction in the last year with Google announcing AutoML ~\cite{automl,zoph2016neural,baker2016designing}.
In situ learning from the environment has also been approached from the direction of spiking neural nets (SNN)
~\cite{roggen2003hardware, kasabov2013dynamic, schuman2016evolutionary}.
Recently intel released a SNN based online learning chip Loihi ~\cite{loihi}. IBM's TrueNorth is also a SNN chip.
SNNs have however not managed to demonstrate accuracy across complex learning tasks.

{\bf DNN Acceleration.} 
Hardware acceleration of neural networks is a hot research topic with a lot of architecture choices ~\cite{chen2014diannao, dadiannao, shidiannao, zhang2015optimizing, chen2016eyeriss, parashar2017scnn, albericio2016cnvlutin, han2016eie, kung2016dynamic} and silicon implementations~\cite{isscc_2016_chen_eyeriss, sim201614, desoli201714, moons2017envision}.
These accelerators can replace \adam for inference, when genes are used to represent layers in MLPs as discussed above. However, \eve remains non-replaceable as there is no hardware platform for efficient evolution in the present to the best of our knowledge.

\vspace{-2mm}
\section{Acknowlegements}
\label{sec:ack}
\vspace{-2mm}
We would like to thank Hyoukjun Kwon, Yu-Hsin Chen, Neal Crago, Michael Pellauer, Sudhakar Yalamanchili, Suvinay Subramanian, and the anonymous reviewers for their insights on the paper.

\vspace{-2mm}
\section{Conclusion}
\label{sec:conclusion}
\vspace{-2mm}

This work presents \system, a system to perform automating NN topology and weight generation completely in hardware.
We first characterize a NE algorithm called NEAT, and identify massive opportunities for parallelism.
Exploiting this, we design two accelerators, EvE and ADAM to accelerate the learning and inference components of NEAT in hardware.
We also perform optimized CPU and GPU implementations and find that they suffer from high power consumption (as expected) and 
low performance due to extensive memory copies.
We believe that this work takes a first key step in co-optimizing NE algorithms and hardware, and opens up lots of exciting avenues for future research.


\bibliographystyle{ieeetr}
\bibliography{neuroevol_ref,new}

\end{document}